# Design and Implementation of a Tool for Extracting Uzbek Syllables

Ulugbek I. Salaev
*Department of Information Technologies*
Urgench State University
Urgench, Uzbekistan
0000-0003-3020-7099

Elmurod R. Kuriyozov
*Department of Information Technologies*
Urgench State University
Urgench, Uzbekistan
0000-0003-1702-1222

Gayrat R. Matlatipov
*Department of Information Technologies*
Urgench State University
Urgench, Uzbekistan
0000-0002-0920-5278

*Abstract*—The accurate syllabification of words plays a vital role in various Natural Language Processing applications. Syllabification is a versatile linguistic tool with applications in linguistic research, language technology, education, and various fields where understanding and processing language is essential. In this paper, we present a comprehensive approach to syllabification for the Uzbek language, including rule-based techniques and machine learning algorithms. Our rule-based approach utilizes advanced methods for dividing words into syllables, generating hyphenations for line breaks and count of syllables. Additionally, we collected a dataset for evaluating and training using machine learning algorithms comprising word-syllable mappings, hyphenations, and syllable counts to predict syllable counts as well as for the evaluation of the proposed model. Our results demonstrate the effectiveness and efficiency of both approaches in achieving accurate syllabification. The results of our experiments show that both approaches achieved a high level of accuracy, exceeding 99%. This study provides valuable insights and recommendations for future research on syllabification and related areas in not only the Uzbek language itself, but also in other closely-related Turkic languages with low-resource factor.

## I. Introduction

Syllabification is the process of dividing words into syllables, which are the units of sound that make up words. Syllabification, hyphenation, and the count of syllables are important in Natural Language Processing (NLP) because they provide important linguistic features that can be used in various natural languages processing tasks, such as text-to-speech synthesis, automatic speech recognition, and text preprocessing. Syllabification is also useful in a range of NLP tasks, such as automatic speech recognition, speech synthesis, and language modelling. By dividing words into syllables, NLP models can better understand the phonetic structure of language, which is useful in tasks such as speech segmentation, prosody modelling, and speech emotion recognition. Additionally, syllabification can help improve the accuracy of NLP models by providing more precise information about the pronunciation of words.

In the Uzbek language, syllabification can be particularly challenging due to the complex syllable structure and the presence of vowel harmony, which is a common feature of the Turkic language family. For Uzbek, syllabification also takes into account vowel sounds. A syllable can either begin with a vowel, meaning it is unobstructed, or it can start with a consonant, making it obstructed. Syllables never begin with two consonants; they always end with just one consonant. Additionally, a syllable can either end with a vowel, making it open, or it can end with a consonant, making it closed.

**Uzbek language** (native: Oʻzbek tili) belongs to the Karluk branch of the Turkic language family. It is characterized as a low-resource and highly-agglutinative language, featuring null-subject and null-gender characteristics. Uzbek is the official language of Uzbekistan and ranks as the second most widely spoken language among Turkic languages. Due to its complex agglutinative nature, Uzbek poses challenges for various language processing tasks, including syllabification. Currently, the Uzbek language utilizes two distinct alphabetic scripts: Latin and Cyrillic. Historically, the Cyrillic alphabet was predominant until its replacement with the Latin script in 1993, which is now the official alphabet. Interestingly, both Cyrillic and Latin scripts are equally popular in various domains of written language, including law, books, web content, media, and more. As a result, NLP research and tools for the Uzbek language are predominantly designed to accommodate both scripts while developing language resources and models. Uzbek has seen considerable growth in the field of NLP research, including morphological analyser [1], [2], semantic evaluation [3], and Uzbek WordNet [4].

In this project, we proposed to develop the methodology of rule-based syllabification including dividing syllables, hyphenation and identifying the count of syllables. Additionally, we trained with Machine Learning (ML) models to predict the count of syllables at the word level for an experiment.

Considering the existence of two distinctive alphabets currently use in the Uzbek language, we focused on the methodology to perform the task of rule-based syllabification for those two alphabets, based on linguistic resources for Uzbek [5].

There are few works on the syllabification task studies of the Uzbek language, besides none of the earlier works offering neither an open-source code nor an Application Programming Interface (API) for integration with other tools. Due to the fact that there is no complete project for syllabification tasks for the Uzbek language, we have to create projects including hyphenation (text-alignment in typing programs), dividing words into syllables, and determining the number of syllables in word functions. Taking this into account, we have created a project that provides both Cyrillic and Latin alphabets for Uzbek, web tools, API and open source-code library for the above-mentioned syllabification tasks. In this paper, we also present a publicly available Python code for research integration, together with a web-based tool that includes an API, which is, to our knowledge, the first-ever syllabification tool including both Uzbek scripts.

## II. RELATED WORK

One of the very early mentions of syllabification was raised by Marchand, Adsett and Damper [6], the automatic syllabification of words poses a significant challenge due to the elusive nature of defining syllables precisely. This task holds importance in various applications such as word modeling for concatenative synthesis and automatic speech recognition. Two primary approaches exist for automatic syllabification: rule-based and data-driven. The rule-based method embodies specific theoretical positions on syllables, while the data-driven paradigm derives new syllabifications from correctly-syllabified examples. This paper compares the performance of these two approaches, considering the difficulty of establishing a definitive "gold standard" corpus for evaluation purposes. To address this challenge, three lexical databases of pre-syllabified words were utilized, allowing for a comprehensive evaluation. The results consistently demonstrate that the data-driven techniques, including a look-up procedure, an exemplar-based generalization technique, and Syllabification by Analogy (SbA), outperform the rule-based system in terms of word and juncture accuracies by a significant margin. Among these techniques, SbA yields the best results.

The research work by Saimaiti and Feng[7] presents a syllabification algorithm designed specifically for written Uyghur, a Turkic language. The algorithm utilizes a set of simple and precise syllabification rules implemented through an abstract computational structure. In the experiment conducted on a random sample, the algorithm demonstrates high accuracy, achieving 98.7% accuracy on word tokens, 99.2% on word types, and 99.1% on syllable accuracy. Additionally, the paper analyzes various aspects of the Uyghur syllable structure based on statistical results obtained from syllabifying about 30K words in the Uyghur Dictionary and more than 2,5M words from a corpus. The findings highlight the effectiveness of a corpus-based approach in computational phonology, producing favourable results for syllabification in Uyghur.

Later modern approaches to syllabification include Machine Learning (ML) models. The work by Krantz et. al.[8] presents an approach to syllabification by treating it as a sequence labelling task. The proposed syllabifier employs an LSTM and convolutional network architecture and leverages a linear chain conditional random field (CRF) for sequence decoding. The model achieves remarkable accuracies surpassing existing approaches for various languages, including Dutch, Italian, French, and Basque, and approaches the best-reported accuracy for English. The paper also provides a freely-available dataset of transcribed syllabification data for evaluation.

The paper by Dinu et. al.[9] highlights the significance of syllable boundary identification in text-to-speech applications and presents a sequence tagging approach using character n-grams with end-of-word marking. The study compares this approach with support vector machines and rule-based methods, demonstrating the effectiveness of the sequence tagging approach in achieving high accuracy in orthographic syllabication tasks specific to Romanian.

The paper by Johnson and Kang[10] focuses on the automatic detection of prominent syllables in Brazil's prosodic intonation model. It investigates the performance of five machine learning classifiers and seven sets of features, including pitch, intensity, and duration, either individually or in combination. The study demonstrates that utilizing pitch, intensity, and duration as features leads to optimal results. The findings highlight that bagging an ensemble of decision tree learners achieves the best performance in terms of accuracy, F−measure, and Cohen's kappa coefficient. The model developed in this study outperforms existing automatic detection software and even human transcription experts in prosody. This research significantly contributes to the field of syllable detection and establishes a new benchmark for detecting prominent syllables in Brazil's prosodic intonation model.

Text-to-speech (TTS) systems play a crucial role in enabling human-machine interaction, and the quality of TTS output is largely determined by the accuracy of the text processing component, particularly word syllabification. In this paper [11], the authors focus on improving the syllabification of the Malay language, by proposing a novel approach and evaluating its performance against three existing methods. The authors collected a dataset of 25K words from Malay language online newspaper articles and Wiktionary Open Content Dictionary and achieved the lowest Word Error Rate (WER) of 2.61% with an accuracy rate of 97.39% with their proposed approach.

Regarding the syllabification works for related languages from the Turkic family, to the knowledge of the authors, there has been very few research, except for Turkish and Uyghur.

In the field of syllabification analysis in the Uzbek language, research works [12], [13] refer to their text-to-speech synthesis project. The Uzbek language has complex syllable structures. The proposed algorithm based on concatenative methods in this paper offers a promising solution, with the development of a comprehensive electronic dictionary of 31K words and their syllables. Based on the dataset, was established 9 patterns of syllable structures. It is explored rule-based approaches to syllabification with a used dataset which does not include inflectional words, but there is a need for more efficient and accurate methods.

Among some other NLP work that has been done on the low-resource Uzbek language so far, there is a morphological analyzer [14], and an Uzbek transliteration project between Uzbek alphabets [15].

## III. METHODOLOGY

In this section, we present all the detailed description of the UzSyllable tool creation, together with how we collected the source data and the rule-based algorithm behind the syllabification tool.

The methodology for syllabification in the context of the UzSyllable tool encompasses a systematic process that transforms raw input text into segmented syllables. The process can be seen in Fig *1*.

A syllable is a sound or combination of sounds uttered by a single breath. For example, *o-na* (mother), *bo-la* (child). A syllable is a phonological unit of a speech stream that is larger than a sound and smaller than a word (sometimes equal to one word), and the field of its study is called syllabification. A syllable cannot be formed without a vowel, so the vowel sound is the centre of the syllable — it draws the consonants to itself and forms a phonetic unit (syllable) that is pronounced with one air stroke.

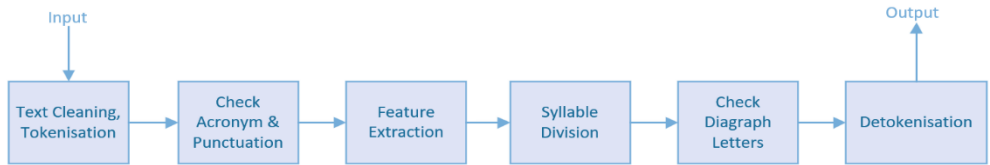

Fig 1. Pipeline for Rule-based approach.

This process involves the following sequential steps:

- **Text Cleaning and Tokenization:** The process commences with the input text, to facilitate accurate processing, the input text undergoes thorough cleaning to remove extraneous characters or artifacts. Especially, the Uzbek texts in Latin alphabet have inconsistency issues with the digraph[1] letters such as *o'* and *g'*, as well as the glottal stop letter ('), which often is written with different characters from the keyboard, mostly with single apostrophes. A special cleaning step has been added to check for these inconsistencies and replace them into correct forms if necessary, also assuring that both characters in the digraphs stay non-separated. Subsequently, tokenization is applied to split the cleaned text into individual words, forming the basis for subsequent analysis.

- **Acronym and Symbol Handling:** Special attention is given to the identification of acronyms and special symbols within the tokenized words. These elements are isolated and marked for preservation throughout the syllabification process.

- **Feature Extraction and Syllable Division:** The core of the syllabification methodology lies in feature extraction and syllable division. This step involves the application of linguistic rules and patterns specific to the Uzbek language to determine appropriate syllable boundaries within each word. This step solely relies on linguistic rules obtained from the Uzbek linguistics book [16] and the explanatory dictionary for Uzbek [17].

- **Correction of Digraph Letters:** The methodology further addresses the intricacies of digraph letters that may span syllable boundaries. A correction step ensures that digraphs are appropriately handled to maintain the integrity of syllable division.

- **Detokenization (output):** Following successful syllable division, the detokenization process merges the segmented words back into a cohesive text, wherein syllables are visually represented through hyphens to distinguish the syllable boundaries.

Based on this methodology, we also created a Python library created for this work (https://github.com/UlugbekSalaev/UzSyllable), which is openly accessible, and also can be easily installed, using the following command that is popular in the Python community:

*pip install UzSyllable*

Moreover, this library was used to create an openly-accessible API, which the details can be seen in Fig . There is also an openly-available web-based syllabification tool (https://nlp.urdu.uz/?menu=syllables), which the user interface can be seen in .

Fig 2. API of the created syllabification tool.

Fig 3. Web interface of the created syllabification tool.

## IV. EXPERIMENTS AND RESULTS

This section outlines a series of comprehensive experiments designed to evaluate the capabilities and efficacy of the UzSyllable tool. These experiments encompass a range of activities, including meticulous data collection, the execution of the tool on extracted words, and the application of diverse machine learning algorithms for detailed analyses of syllable prediction accuracy and performance.

---

[1] Digraph letters are the ones that are made of two characters to form one letter.

### A. Dataset Collection

The foundation of our experimental evaluation begins with the meticulous assembly of a robust dataset, crucial for assessing the UzSyllable tool's performance. Our dataset acquisition process involved the manual extraction of words and their corresponding syllabifications from the "The Uzbek Dictionary" hardcopy book. This arduous effort yielded an extensive collection of 78,541 words with their associated syllable divisions in the Latin script, while 13,571 words were similarly extracted for the Cyrillic script.

The extracted dataset consists of the original word, its correct syllabification, hyphenation variants, as well as the syllable counts for each script. A detailed description of the sample view of the extracted dataset can be seen in TABLE I.

### B. Experiments

Initially intended as a direct comparison of syllabification results with existing tools, due to the absence of suitable open-source alternatives. Consequently, we redirected our focus towards predicting syllable counts, an equally informative task that allowed us to quantitatively measure the accuracy and reliability of the UzSyllable tool.

TABLE I. SIMPLE VIEW OF THE DATASET (LATIN, CYRILLIC)

| Word | Syllables | Hyphenation | Syllable count |
|---|---|---|---|
| abadiy | a-ba-diy | aba-diy | 3 |
| abadiyat | a-ba-di-yat | aba-diyat abadi-yat | 4 |
| abobil | a-bo-bil | abo-bil | 3 |
| arabcha | a-rab-cha | arab-cha | 3 |
| adovatli | a-do-vat-li | ado-vatli adovat-li | 4 |
| bug'latkich | bug'-lat-kich | bug'-latkich bug'lat-kich | 3 |
| chaldirtirish | chal-dir-ti-rish | chal-dirtirish chaldir-tirish chaldirti-rish | 4 |
| dahshatli | dah-shat-li | dah-shatli dahshat-li | 3 |
| keksaygan | kek-say-gan | kek-saygan keksay-gan | 3 |
| kichkintoy | kich-kin-toy | kich-kintoy kichkin-toy | 3 |
| qadoqlatish | qa-doq-la-tish | qa-doqlatish qadoq-latish qadoqla-tish | 4 |

To achieve this, we leveraged a selection of machine learning algorithms, each designed to capture distinct patterns and relationships within the data. These algorithms included:

- **Decision Tree Regressor (DTRegressor):** Employing decision trees, this algorithm partitions the dataset into subsets based on attribute values, enabling accurate predictions by analyzing the hierarchy of conditions.
- **K-Nearest Neighbors (KNN):** KNN predicts syllable counts based on the similarity between input data and its k-nearest neighbours, where the output is determined by the majority class of those neighbours.
- **Random Forest Regressor (RFRegressor):** Utilizing an ensemble of decision trees, this algorithm enhances predictive accuracy by aggregating multiple individual predictions.
- **Support Vector Machine (SVM):** SVM constructs hyperplanes to classify data points into different classes. In the context of syllable count prediction, it seeks to find an optimal hyperplane that best separates the input features.
- **Multi-Layer Perceptron Regressor (MLPRegressor):** This neural network architecture comprises multiple layers of interconnected nodes, enabling the algorithm to learn complex relationships and patterns in the dataset.
- **Recurrent Neural Network (RNN):** Specializing in sequential data, RNNs are well-suited for predicting syllable counts by considering the sequential nature of syllable structures.

These machine learning algorithms were trained on the dataset consisting of extracted words and their corresponding syllable counts, enabling them to learn and generalize from the data patterns. The subsequent evaluation phase aimed to quantify the predictive accuracy of each algorithm and compare their performance in predicting syllable counts for Uzbek words. The following section presents the intricate details of our experimental results, shedding light on the effectiveness of the UzSyllable tool and the insights gained from our analyses.

We applied ML algorithms with 5-fold cross-validation for the evaluation, the standard deviation of all the results was found to be lower than 0,03. This indicates that the performance of the ML models was relatively consistent across the cross-validation folds, suggesting a stable and reliable performance in the given task.

TABLE II. MICRO-AVERAGED F1 SCORES OF SYLLABIFICATION METHOD IN A WORD LEVEL FOR DATASETS (LATIN, CYRILLIC).

| Classifiers | Latin | Cyrillic |
|---|---|---|
| UzSyllable tool | 0.9872 | 0.9443 |
| Decision Tree Regressor | 0.9935 | 0.9561 |
| K-Nearest Neighbors (KNN) | 0.9989 | 0.9488 |
| Random Forest Regressor | 0.9998 | 0.9536 |
| Support Vector Machine (SVM) | 0.9981 | 0.9606 |
| Multi-Layer Perceptron Regressor | 0.9998 | 0.9537 |
| Recurrent Neural Network | 0.9741 | 0.9873 |

*C. Results*

For the assessment of overall experimental performance, we opted for the micro-averaged F1-score as the chosen evaluation metric. The comparison of the results of our UzSyllable tool with the results of machine-learning algorithms is reported in TABLE II.

The machine learning-based classifiers, such as DTRegressor, SVM, MLPRegressor, and RNN, slightly outperformed our rule-based approach on both Latin and Cyrillic datasets, achieving near-perfect or higher F1 scores. This suggests that machine learning algorithms offer improved accuracy and effectiveness in syllabification tasks compared to traditional rule-based methods for the given datasets.

## V. DISCUSSION

While the created tool proves to be practical for syllabification purposes, certain cases require further consideration and ongoing improvement:

• The dataset is limited to root and derivational words. Considering the strong agglutinative nature of the Uzbek language, a significant portion of words in the text consists of inflectional forms. However, the reason why the test dataset does not include the inflectional forms of the Uzbek words is because the created rule-based syllabification tool already covers all the possible inflectional forms in the text;

• Loanwords and proper nouns adopted from other languages might not produce expected output, thus we have to store exceptional cases in a database;

• The syllable division method yielded incorrect results for derived words such as *kadastrlash*, *magistrlik*, *silindrli*. However, the method provided correct results for their root forms (*kadastr*, *magistr*, *silindr*). It was observed that when Uzbek language affixes are added to borrowed words, the resulting syllable division does not align with the general rules of the Uzbek language.

• We dealt with legal text properly written in Uzbek language, which is not always the case with user-generated text. Especially, there is a big deal of inconsistency in writing *o'* and *g'* letters in the currently official alphabet due to the use of apostrophe, which comes in many ways, such as *o',o',o'*, and *g',g',g'* forms respectively;

TABLE III presents the discrepancies between the rule-based model's output and the correct syllables for selected words, showcasing scenarios where the rule-based approach may not be sufficient for precise syllable division.

## VI. CONCLUSION AND FUTURE WORK

In this paper, we introduced a rule-based algorithm for the task of syllabification of texts in the Uzbek language, encompassing both the Latin and Cyrillic alphabets that are equally in use. Besides, a Python library, web-based user interface, as well as details of an accessible API for the created tool - UzSyllable have been presented. A comprehensive dataset was created to evaluate the performance of the proposed model, and various machine learning algorithms were applied to the dataset for experiments.

Moreover, cases of exceptional situations related to the syllabification according to one of the general mistakes type that occur in loanwords that deviate from the spelling conventions of the language. In future work, we plan to enhance the UzSyllable tool by incorporating machine learning techniques into the tool to further improve syllabification accuracy and extend its capabilities to handle diverse dialects and language variations within the Uzbek language.

TABLE III. SOME EXAMPLES FROM THE MODEL'S PREDICTION USING DATASET WHERE RULE-BASED MODEL DOES NOT APPLY.

| Word | Correct Syllables | Predicted Syllables |
|---|---|---|
| **abstrakt** | ab-strakt | abst-rakt |
| **agglyutinativ** | a-gglyu-ti-na-tiv | aggl-yu-ti-na-tiv |
| **alebastrchi** | a-le-ba-strchi | a-le-bastr-chi |
| **ansamblchi** | an-sa-mblchi | an-sambl-chi |
| **avtotransplantatsiya** | av-tot-ra-nsplan-tat-si-ya | av-tot-ransp-lan-tat-si-ya |
| **aviaekspress** | a-vi-a-eks-press | a-vi-a-eksp-ress |
| **aviakonstruktor** | a-vi-a-ko-nstruk-tor | a-vi-a-konst-ruk-tor |
| **avstraliyalik** | a-vstra-li-ya-lik | avst-ra-li-ya-lik |
| **bergshtrixlar** | berg-shtrix-lar | bergsht-rix-lar |
| **eksklyuziv** | e-ksklyu-ziv | ekskl-yu-ziv |
| **ekstremizm** | e-kstre-mizm | ekst-re-mizm |
| **elektrlampa** | e-le-ktrlam-pa | e-lektr-lam-pa |
| **industrlash** | in-du-strlash | in-dustr-lash |
| **inflyatsiya** | i-nflyat-si-ya | infl-yat-si-ya |
| **instruksiya** | i-nstruk-si-ya | inst-ruk-si-ya |
| **kadastrlash** | ka-da-strlash | ka-dastr-lash |
| **magistrlik** | ma-gi-strlik | ma-gistr-lik |
| **silindrli** | si-lin-drli | si-lindr-li |

ACKNOWLEDGEMENT

This research work was funded by the REP-25112021/113 - "UzUDT: Universal Dependencies Treebank and parser for natural language processing on the Uzbek language" subproject funded by The World Bank project "Modernizing Uzbekistan national innovation system" hosted by The National University of Uzbekistan named after Mirzo Ulugbek.